\documentclass[letterpaper, 10 pt, conference]{ieeeconf}  % Comment this line out if you need a4paper

\IEEEoverridecommandlockouts                              % This command is only needed if 
\usepackage{graphics} % for pdf, bitmapped graphics files
\usepackage{epsfig} % for postscript graphics files
\usepackage{mathptmx} % assumes new font selection scheme installed
\usepackage{times} % assumes new font selection scheme installed
\usepackage{amsmath} % assumes amsmath package installed
\usepackage{amssymb}  % assumes amsmath package installed
\usepackage{cite}
\usepackage[style=base]{subcaption}
\usepackage[binary-units=true]{siunitx}
\usepackage[ruled,lined,linesnumbered]{algorithm2e}
\let\oldnl\nl% Store \nl in \oldnl
\newcommand{\nonl}{\renewcommand{\nl}{\let\nl\oldnl}}
\graphicspath{{figures/}}
\DeclareCaptionLabelSeparator{periodspace}{.\quad}
\title{\LARGE \bf
Bio-mimetic Adaptive Force/Position Control Using Fractal Impedance 
}

\author{Carlo Tiseo$^{1}$, Wolfgang Merkt$^{1,2}$, Keyhan Kouhkiloui Babarahmati$^{1}$,  Wouter Wolfslag$^{1}$, \\Sethu Vijayakumar$^{1}$ and Michael Mistry$^{1}$% <-this % stops a space
\thanks{This work has been supported by the the Engineering and Physical Sciences Research Council (EPSRC) UK RAI Hub ORCA (EP/R026173/1).}% <-this % stops a space
\thanks{$^{1}$ Edinburgh Centre of Robotics, Institute of Action, Perception, and Behaviour,  School of Informatics, University of Edinburgh, Edinburgh, Scotland, UK.
        {\tt\small ctiseo@ed.ac.uk}}%
\thanks{$^{2}$ Oxford Robotics Institute, University of Oxford, Oxford, England, UK.}%
}

\begin{document}
\thispagestyle{empty}
\fbox{
\parbox{\textwidth}{
© 2020 IEEE.  Personal use of this material is permitted.  Permission from IEEE must be obtained for all other uses, in any current or future media, including reprinting/republishing this material for advertising or promotional purposes, creating new collective works, for resale or redistribution to servers or lists, or reuse of any copyrighted component of this work in other works.}}
\newpage
\maketitle
\thispagestyle{empty}
\pagestyle{empty}

%%%%%%%%%%%%%%%%%%%%%%%%%%%%%%%%%%%%%%%%%%%%%%%%%%%%%%%%%%%%%%%%%%%%%%%%%%%%%%%%
\begin{abstract}

The ability of animals to interact with complex dynamics is unmatched in robots. Especially important to the interaction performances is the online adaptation of body dynamics, which can be modelled as an impedance behaviour. However, variable impedance control still continues to be a challenge in the current control frameworks due to the difficulties of retaining stability when adapting the controller gains. The fractal impedance controller has recently been proposed to solve this issue. However, it still has limitations such as sudden jumps in force when it starts to converge to the desired position and the lack of a force feedback loop. In this manuscript, two improvements are made to the control framework to solve these limitations. The force discontinuity has been addressed introducing a modulation of the impedance via a virtual antagonist that modulates the output force. The force tracking has been modelled after the parallel force/position controller architecture. In contrast to traditional methods, the fractal impedance controller enables the implementation of a search algorithm on the force feedback to adapt its behaviour to the external environment instead of on relying on \textit{a priori} knowledge of the external dynamics. Preliminary simulation results presented in this paper show the feasibility of the proposed approach, and it allows to evaluate the trade-off that needs to be made when relying on the proposed controller for interaction. In conclusion, the proposed method mimics the behaviour of an agonist/antagonist system adapting to unknown external dynamics, and it may find application in computational neuroscience, haptics, and interaction control.

\end{abstract}

%%%%%%%%%%%%%%%%%%%%%%%%%%%%%%%%%%%%%%%%%%%%%%%%%%%%%%%%%%%%%%%%%%%%%%%%%%%%%%%%
\section{Introduction}
The dynamic dexterity of animals is far beyond the current capabilities of artificial systems, despite their theoretical limitation in information transmission and processing \cite{Hogan2012, Hogan2013,tiseo2018BB,tiseo2018}. An important limitation in state-of-the-art control frameworks is the reduced performance and reliability when dealing with highly variable dynamic interaction (e.g., contacts). Here, a major limitation is the dependence of their stability on an accurate interaction dynamics model; which inherently is not easy to track with the unpredictability of real-world interaction scenarios \cite{angelini2019,braun2013robots}. Therefore, the identification of a method to overcome such limitation is likely to have a significant impact on the robustness of legged robots such as quadrupeds and humanoids, haptic technologies, exoskeletons, and rehabilitation systems. 

One of the theories on how nature has overcome these biological limitations is based on the theory that the body motor control acts as a hierarchical architecture of semi-autonomous controllers \cite{Ahn2012, tiseo2018}. In this type of architecture, each layer is accountable for the stability while accurately executing the behaviour directed from the higher level controller, while at the same time coordinating and verifying the behaviour of lower level controllers. This can be observed starting from the muscular level where each muscle contains structures that can modulate its mechanical impedance \cite{shadmehr1991,franklin2008,ganesh2010,Mussa-Ivaldi1985}. Moving to a higher level, motor synergies can be identified that co-activate muscles, often across multiple joints, to produce desired body motions \cite{tiseo2018,Hogan2013,Hogan2012,Ahn2012,shadmehr2017}. The motor synergies themselves are identified at different hierarchies starting from a single agonist/antagonist behaviour to more complex stereotyped movements that are implemented in the coordination of complex actions, such as balance \cite{tiseo2018}. In summary, there is a multitude of studies indicating that the motor control modulates the mechanical impedance of the human body to adapt to the task requirements.

\begin{figure}[t]
    \centering
    \includegraphics[width=\linewidth]{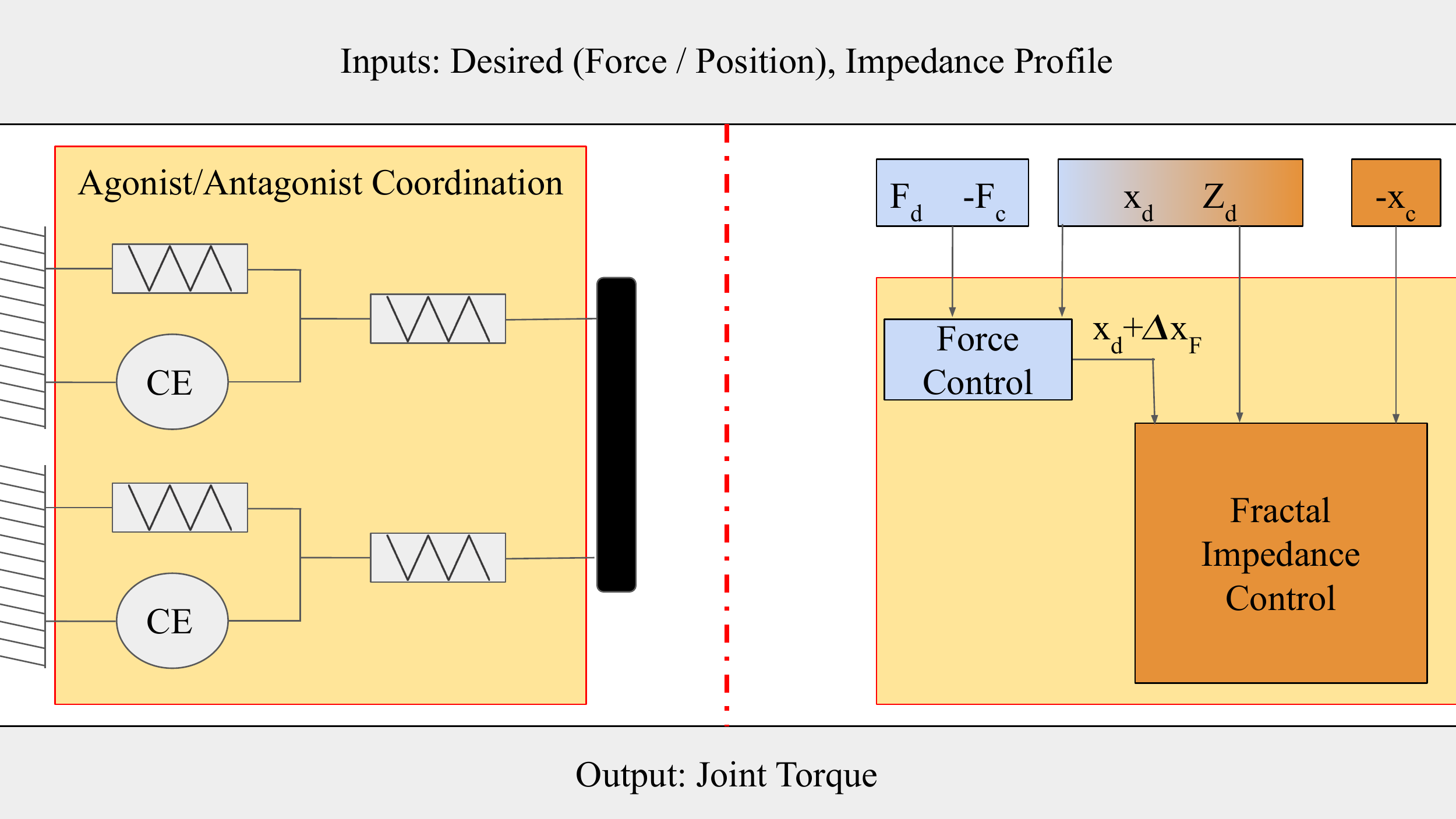}
    \caption{Agonist/Antagonist muscles perform synergistic actions to control the joints' movements. The experimental evidence also confirmed that their impedance is adjusted based on different tasks, and can be modulated online by the contracting elements (CT) to adapt to external perturbations. The proposed integration of a parallel force control with the fractal impedance controller aims to reproduce similar capabilities. Our controller requires as inputs desired force (F$_\text{d}$), desired position (x$_\text{d}$), desired impedance profile (Z$_\text{d}$), and the feedback of the current position (x$_\text{c}$) and current force (F$_\text{c}$).}
    \label{fig:00}
\end{figure}

\begin{figure}[!tb]
    \centering
    \begin{subfigure}[b]{0.9\linewidth}
    \centering
    \includegraphics[width=\linewidth]{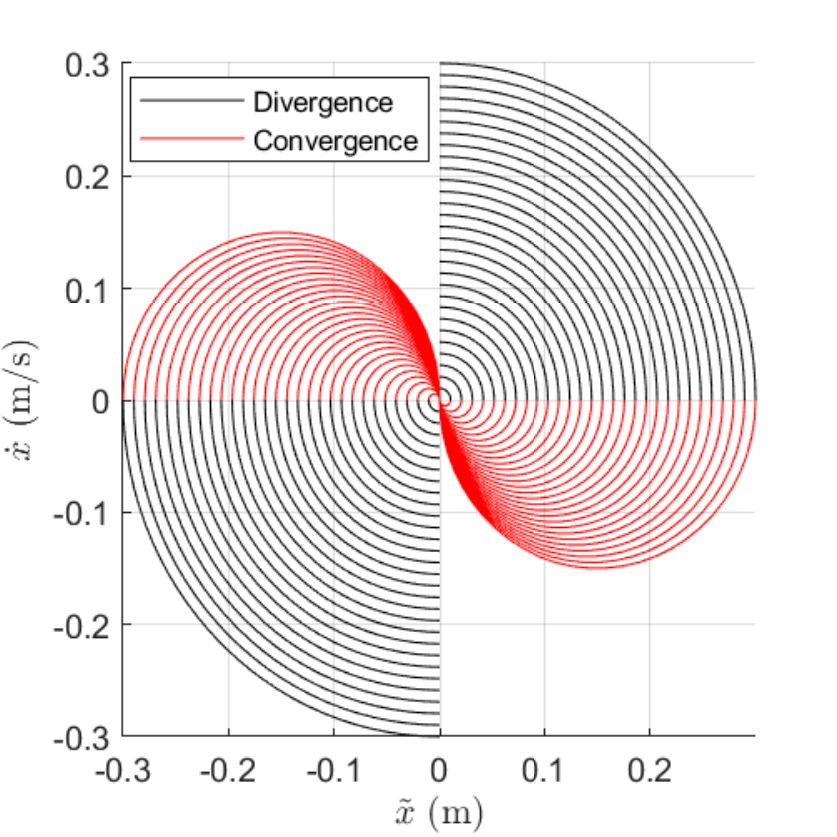}
    \caption{\label{fig:0a}}
     \end{subfigure}%
     \\
   \begin{subfigure}[b]{0.9\linewidth}
    \centering
    \includegraphics[width=\linewidth]{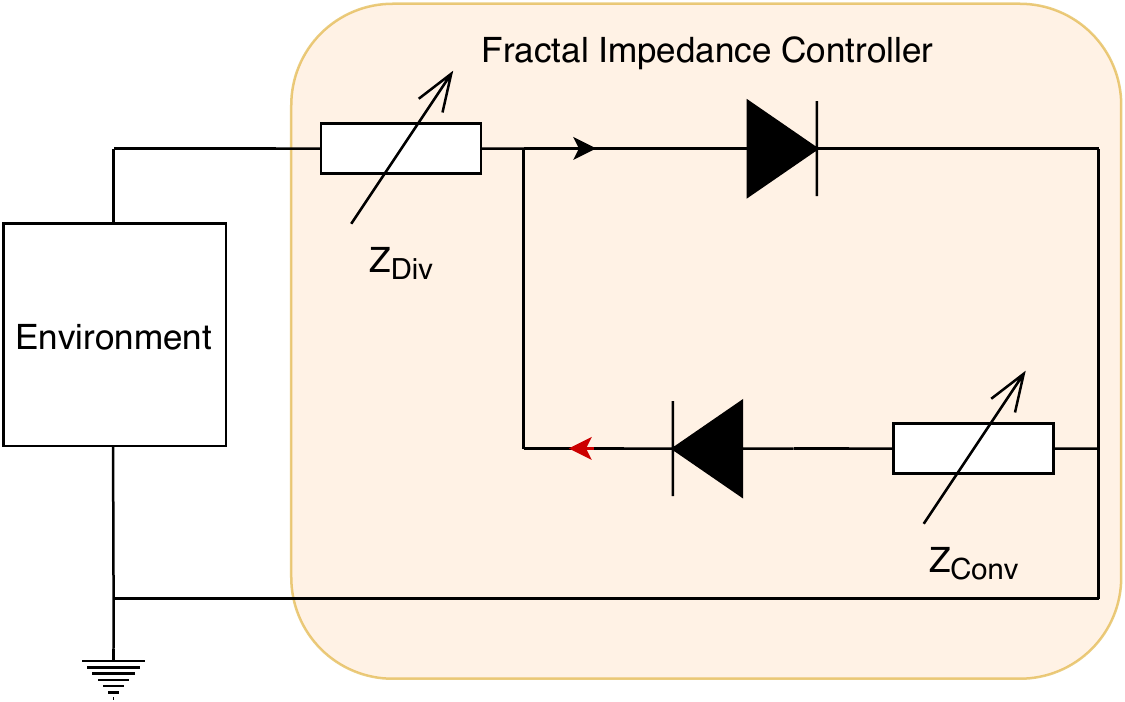}
    \caption{\label{fig:0b}}
    \end{subfigure}%
    \caption{(a) Phase portrait of the fractal attractor. The trajectories that reach a maximum displacement ($\tilde{x}_{\text{Max}}$) greater than 0.3 have been omitted in the plot. (b) The controller behaves like an electrical circuit which has two parallel lines with diodes oriented in opposite directions---this enables to bypass $\text{Z}_{Conv}$ during divergence. The controller stability is achieved guaranteeing the conservation of energy while switching from divergence to convergence.}
    \label{fig:0}
\end{figure}

\begin{figure*}[ht]
    \centering
    \begin{subfigure}[b]{0.5\linewidth}
        \centering 
        \includegraphics[width=0.9\linewidth]{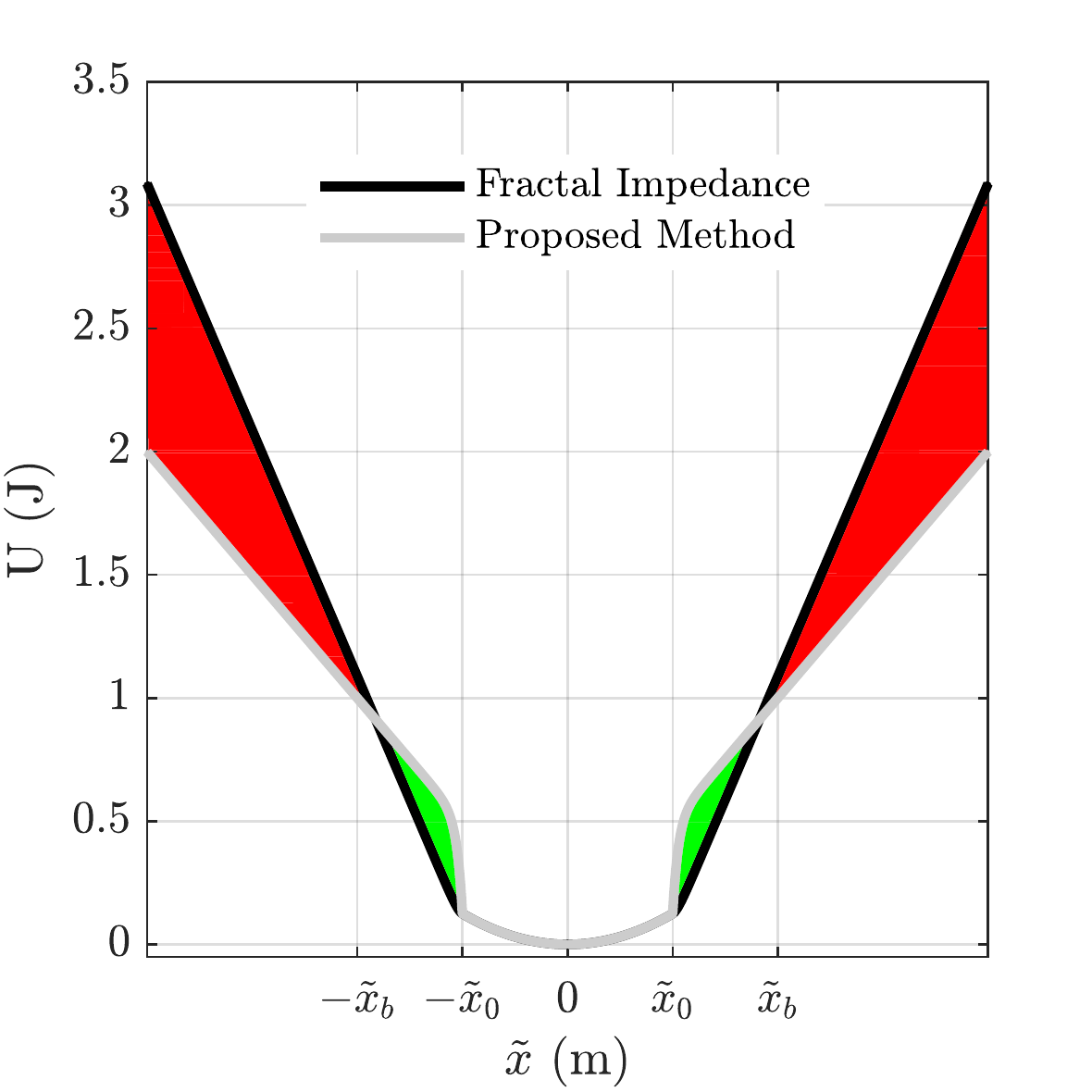}
        \caption{}\label{fig:1a}
    \end{subfigure}%
    \begin{subfigure}[b]{0.5\linewidth}
        \centering 
         \includegraphics[width=0.9\linewidth]{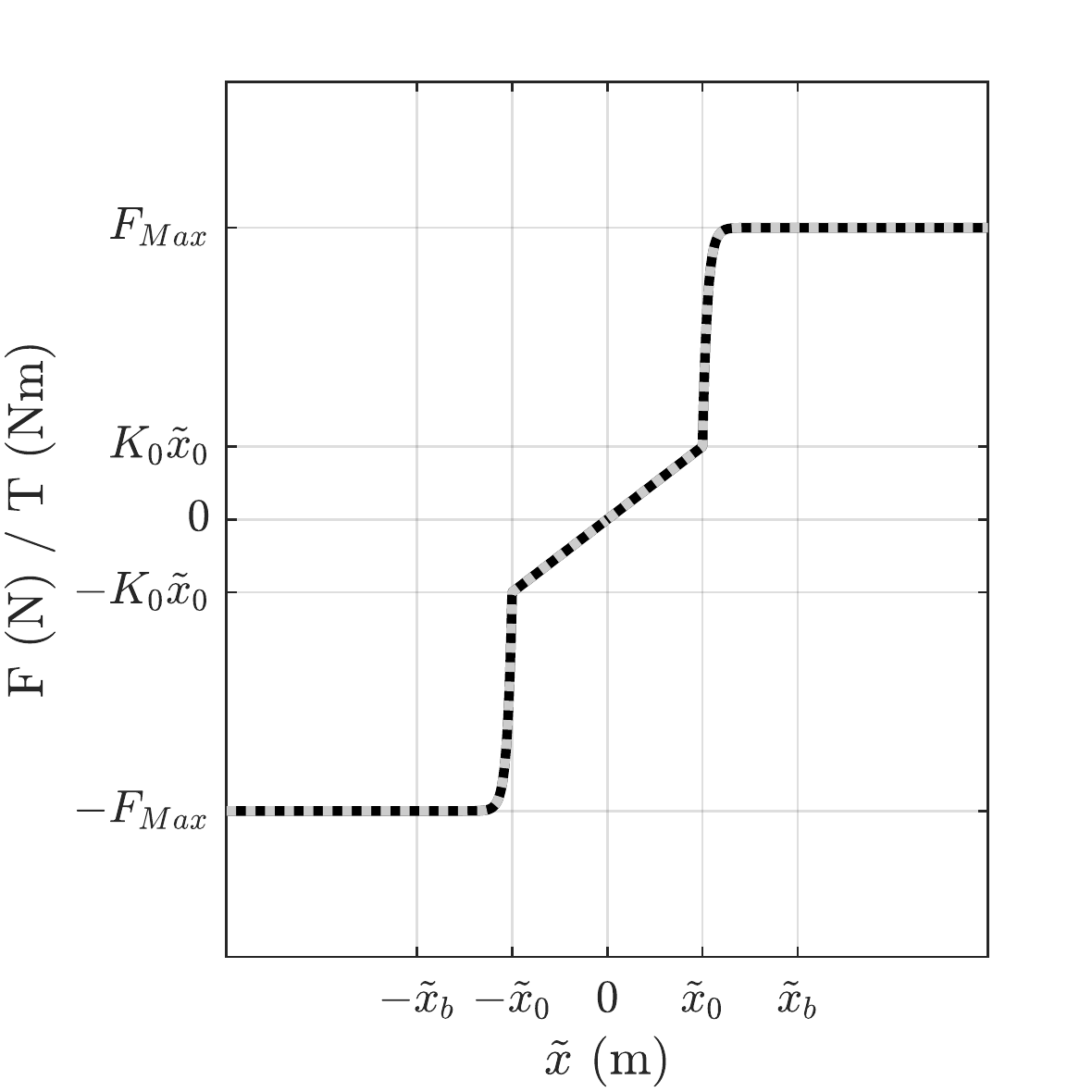}
        \caption{}\label{fig:1b}
    \end{subfigure}
    \caption{(a) The difference between the energy released in the environment from the fractal impedance and the proposed method for obtaining smooth force transition during inversion. As a consequence, an increase in the energy released into the environment emerges in the interval $(\tilde{x}_\text{0},~\tilde{x}_\text{b})$, indicated in green. On the other hand, there is a reduction of the released energy (in red), once the force saturates when $|\tilde{x}|$ is greater than $\tilde{x}_{\text{b}}$. (b) The forces/torques generated at the switching point by the two impedance elements are perfectly superimposed when using the proposed method.}\label{fig:1}
\end{figure*}

The identification of the ability of motor control to modulate the body impedance lead to theories based on the Port-Hamiltonian framework \cite{hogan2013Net}. However, their integration in a complex hierarchical architecture of semi-autonomous controllers is still an open issue. Furthermore, despite interaction control based on the Port-Hamiltonian approach having been proposed in the forms of impedance and admittance control, there are still issues when dealing with variable impedance and the interaction with complex external dynamics. 

Impedance and admittance control differ on the methodology used to decode the physical information exchanged from the system with the environment \cite{hogan2013Net}. Impedance control transforms an incoming flow (i.e., velocity) into the desired effort (i.e., force), while admittance control transforms the incoming effort into the desired kinematics \cite{hogan2013Net}. The stability of Port-Hamiltonian controllers relies on the ability to track their energy exchange, which is challenging when dealing with redundant systems and changing environmental conditions. The approaches for tracking the energy are based on projectors \cite{angelini2019}. This makes these controllers susceptible to changes in the task (e.g., contact, admittance, and impedance changes) as well as singularities which is where these projectors can degenerate numerically \cite{dietrich2015passivation}.

The Fractal Attractor has recently been proposed to solve this issue by using a passive variable impedance. The controller stability is guaranteed by its passivity, which is maintained using the spring energy map of the system state \cite{babarahmati2019,tiseo2020}. The damping component of the controller acts only as an energy sink (i.e., reference velocity set to zero). The controller relies upon an anisotropic variable impedance redistributing the energy absorbed during divergence to converge to the desired state rather than a limit cycle. Other benefits of this controller are that (a) the impedance is defined based on a desired force/displacement behaviour, (b) it can be safely tuned online, (c) multiple controllers can be superimposed without affecting stability, and (d) it can be calibrated to account for the physical limitation of the robot guaranteeing the global stability of the controller. The current formulation has two major limitations which are (1) the lack of a feedback loop on the force, and (2) the lack of force homogeneity when the controllers switch between divergence and convergence phases. 

This manuscript proposes a new impedance profile formulation inspired by the agonist/antagonist configuration of the muscles to solve the force homogeneity issue. We introduce a force-feedback, also occurring in biological actuation, implemented using a reinterpretation of the parallel force/position control force feedback in traditional impedance controllers. The controller is validated in a Simulink (Mathworks Inc, USA) system simulation to evaluate its ability to adapt its behaviour to the interaction with unknown environmental dynamics. 

\section{Fractal Controller}
The Fractal Impedance Controller is based on an attractor that alters its topology between its divergence and convergence phases (Fig.~\ref{fig:0a}) \cite{babarahmati2019}. During divergence, the controller stiffness is described as a spring centred in the desired position. During convergence, the controller acts as a constant spring centred in the mid-point between the desired position and the maximum displacement ($\tilde{x}_{\text{Max}}$) reached during the divergence phase \cite{babarahmati2019}. Such behaviour is obtained with an algorithm that mimics the electrical equivalent as shown in Fig.~\ref{fig:0b}. The behaviour during divergence is determined by $\text{Z}_{Div}$, while an additional impedance $\text{Z}_{Conv}$ is added in series for convergence. Thus, the system behaves like having a total impedance equal to $\text{Z}_{Tot}=\text{Z}_{Div}+\text{Z}_{Conv}$. Furthermore, the fractal impedance controller does not have a damping component, which implies that the controller generates a conservative energy field centred in the desired pose. It shall be noted that a damping component can be introduced without affecting stability as long as the desired velocity is equal to zero \cite{babarahmati2019}.

This approach has been inspired by the agonist/antagonist muscles and the concept of dynamic primitives, which can be seen as libraries of stereotyped basic behaviours that can be superimposed to achieve complex actions \cite{Hogan2012,Hogan2013,Tommasino2017,tiseo2018, schaal2000}. Specifically, the hypothesis was that it should be possible to embed stability into the control algorithm itself identifying a stereotyped behaviour that could scale with the energy accumulated in the system without compromising stability. In other words, is it possible that what we observe as variable impedance in humans is produced by the scaling of intrinsically stable stereotyped embedded systems?

Following such a change of perspective, the problem is transformed into one of how to embed a smooth trajectory into the attractor of the impedance controller itself. The implemented solution, inspired by biology, has been alluded to earlier \cite{babarahmati2019}, enabling the tuning of the desired force/displacement characteristics. However, these characteristics currently come at the cost of a sudden change in force magnitude while starting to converge and the lack of force feedback loop required for tasks such as haptic exploration. 

\subsection{Altering Fractal Impedance Profile for Force Homogeneity}
The fractal attractor proposed in \cite{babarahmati2019,tiseo2020} uses the conservation of energy principle to calculate the desired impedance for convergence, leading to a change of the gradient of energy at the inversion point that results in a sudden change in the exerted force. A solution to this issue may be provided by considering an agonist/antagonist control strategy, in which the two systems are impedance connected in parallel and the total produced effort (i.e., force) can be corrected by modulating the antagonist impedance, which alters the energy output through the port and smooths the force transition. However, the stability of the controllers can be retained only and only if this energy difference is known and, therefore, it can be accounted for in the stability analysis. As detailed in \cite{babarahmati2019}, a non-smooth Lyapunov's candidate for the system must be laterally unbounded and have a bounded finite derivative. To identify the energy profile, a desired output force profile of the spring during divergence has been selected, following the profile presented in \cite{tiseo2020}:

\begin{equation}
    \label{ForceProf}
    F_{\text{K}}=\left\{
    \begin{array}{ll}
      K_0 \tilde{x},  & |\tilde{x}|<\tilde{x}_{0}\\\\
 \text{sgn} (\tilde{x}) (\Delta F (1-e^{-\frac{|\tilde{x}| - \tilde{x}_0}{b}})+& \\
     + K_0 \tilde{x}_0),   & \tilde{x}_{0}\le |\tilde{x}|<\tilde{x}_{b}\\\\
     \text{sgn} (\tilde{x})  F_{\text{Max}},   & \text{Otherwise}
    \end{array}\right.
\end{equation}
where $K_0$ is the constant stiffness, $\Delta F =(F_{\text{Max}} - K_0.\tilde{x}_0)$, $\tilde{x}_{0}$ is the displacement that triggers the non-linear spring to activate, $\tilde{x}_{b}$ indicates the displacement associated with the force saturation, $F_{\text{Max}}$ is the saturation force, $b=(\tilde{x}_\text{b}-\tilde{x}_\text{0})/S$ is the characteristic length of the sigmoidal function. $S=20$ controls the saturation velocity ensuring that the force is at \SI{99.9}{\percent} of $F_{max}$ before reaching $\tilde{x}_{b}$. 
The energy profile associated with this force profile is as follows:
\begin{equation}
    \label{EnergyProf}
    E_{\text{K}}=\left\{
    \begin{array}{ll}
      0.5 K_0 \tilde{x}^2,  & |\tilde{x}|<\tilde{x}_{0}\\\\
      F_{\text{Max}}|\tilde{x}| -&\\
      +F_{\text{Max}}\tilde{x}_0 + (K_0\tilde{x}_0^2)/2 -&\\
      +(1-e^{-\frac{|\tilde{x}| - \tilde{x}_0}{b}})b\Delta F, 
      & \tilde{x}_{0}\le |\tilde{x}|<\tilde{x}_{b}\\\\
      F_{\text{Max}}|\tilde{x}|-&\\
      +F_{\text{Max}}\tilde{x}_0 + (K_0\tilde{x}_0^2)/2 -&\\
      +(1-e^{-\frac{\tilde{x}_b - \tilde{x}_0}{b}})b\Delta F,   & \text{Otherwise}
    \end{array}\right.
\end{equation}
If these profiles are used in the fractal impedance controller algorithm introduced in \cite{babarahmati2019}, they generate a non-smooth transition while switching from divergence to convergence. The force transition can be smoothed using the Algorithm \ref{alg1}. 
However, this algorithm introduces a discrepancy between the accumulated and the released energy (Figure \ref{fig:1a}). The virtual antagonistic muscle is introduced to account for the energy discrepancy in the stability analysis. The Lyapunov proof is included in Appendix A, which updates the proof given in \cite{babarahmati2019}.

\begin{algorithm} 
\SetAlgoLined
  \SetKwData{Left}{left}
  \SetKwData{Up}{up}
  \SetKwFunction{FindCompress}{FindCompress}
  \SetKwInOut{Input}{input}
  \SetKwInOut{Output}{output}
  \Input{Divergence/Convergence, $F_{\text{K}}(\tilde{x})$, $E_{\text{K}}(\tilde{x}_{\text{Max}})$, $\tilde{x}_{\text{Max}}$, $\tilde{x}$, $F_\text{K}(\tilde{x}_{Max})$}
  \Output{$h_e$}
  \eIf{diverging from ${x_{\text{d}}}$}{
     $h_{\text{e}}=F_{\text{K}}(\tilde{x})$\\
  }{
     \eIf{$\tilde{x}<=x_0$}{
     $h_{\text{e}} = F_{AA} (\tilde{x})= \frac{4 E_{\text{K}}(\tilde{x}_{\text{Max}})}{\tilde{x}_{\text{Max}}^2} (\tilde{x}-0.5\tilde{x}_{\text{Max}})$\\
     }{
      $h_{\text{e}} = F_{AA} (\tilde{x})= \frac{2F_\text{K}(\tilde{x}_{Max})}{\tilde{x}_{\text{Max}}}(\tilde{x}-0.5\tilde{x}_{\text{Max}})$
     }
  }
 \caption{Mono-dimensional Fractal Impedance Control}
 \label{alg1}
\end{algorithm}

\subsection{Force Feedback}
A system interacting with an external entity does not only affect the environment but it is also affected by it. Therefore, the feed-forward dynamic interaction enabled by the impedance controller is not sufficient to guarantee the chosen behaviour during interaction. In fact, the environmental impedance will act as a parallel impedance to our controller modifying the resultant behaviour, which implies that we need to adjust $\tilde{x}$ to achieve the desired interaction force. An easy to visualise analogue is considering two springs pushing against each other, with one having a tunable stiffness. It is commonly known that the equilibrium point depends on their relative stiffness. Therefore, if we do not directly alter the exchanged force, we need to adjust the stiffness which also alters the equilibrium point. 

An impedance controller interacting with an infinitely rigid system (i.e., non-deformable) produces the force at the interface predicted by the selected impedance model. However, the displacement ($\tilde{x}$) required to achieve the same force will change when interacting with non-rigid systems, based on their relative stiffness. The parallel Force/Position control was proposed to solve this issue by adjusting the desired position based on a model of the interaction dynamics \cite{Siciliano1996}. The major drawback of that approach was that being applied to a traditional impedance controller, an accurate model and a proper tuning of the impedance controller are essential for retaining the stability. However, the fractal impedance controller does not have such limitations, because the stability condition is embedded into the topology of controller attractor, cf. Fig. \ref{fig:0}. This in turn implies that the availability of an accurate expectation of the environment will only enable a more efficient interaction. Thus, an online force feedback-based haptic exploration (Algorithm \ref{alg2}) can be deployed without introducing any concerns for the controller stability.

\begin{algorithm} 
\SetAlgoLined
  \SetKwData{Left}{left}
  \SetKwData{Up}{up}
  \SetKwFunction{FindCompress}{FindCompress}
  \SetKwInOut{Input}{input}
  \SetKwInOut{Output}{output}

 \Input{$F_{\text{d}}$, $K_0$, $\sigma$ , $F(t-1)$, $\tilde{x}$, $x_\text{d}$, $x_\text{d}^\text{h}(t-1)$  }
 \Output{$x_\text{d}^\text{h}(t)$}
  \BlankLine
  $\delta \tilde{x}_0=\frac{F_{\text{d}}}{K_0}$\\
  $\delta \tilde{x}_0^\text{h}=\sigma\delta \tilde{x}_0$\\
  \eIf{$F(t-1)~=F_\text{d}(t)$}{
      \eIf{$|\tilde{x}|\le |x_\text{d}|$
           }{
                $x_\text{d}^\text{h}(t)=x_\text{d}+\delta \tilde{x}_0$
           }{
                $\Delta x=\frac{(F_\text{d}(t)-F(t-1))}{F_{\text{d}}(t)}\delta \tilde{x}_0^\text{h}$\\
                $x_\text{d}^\text{h}(t)=x_\text{d}^\text{h}(t-1)+\Delta x$\\
            }
    }{
         $x_\text{d}^\text{h}(t)=x_\text{d}^\text{h}(t-1)$
    }
\nonl   where:
\nonl   $t$ is the discrete time variable,\\
\nonl   $F_{\text{d}}$ is the desired force,\\
\nonl   $x_\text{d}$ is the displacement from the reference position that is expected when making contact with the environment,\\
\nonl $\sigma=0.01$ scaling factor for the force scanning resolution.
    
\caption{Haptic Exploration} \label{alg2}
\end{algorithm}

Lastly, it shall be noted that the proposed algorithm is made possible because within the fractal impedance controller the conservative field determined by the stiffness is the only source of energy. Therefore, the only interaction that can alter the topology generated by the controller stiffness is the coupling with an external non-infinite stiffness. On the other hand, the mass and damping component will affect only the path taken for converging to the desired behaviour.   

\section{Experimental Design}
A simulator is a good approach for the preliminary evaluation of the proposed method because it allows direct control of the mechanical properties of the components involved without building a dedicated structure. This allows us to study how changes in properties of the external dynamics affect the controller performance and \textit{vice versa}. The simulations are realised in Simulink (MathWorks, Inc). The solver is the \texttt{ode4} with a constant time-step of \SI{10}{\milli\second} for continuous contact dynamics. The variable step solver \texttt{ode23t} has been used due to issues encountered in solving the model of the switching contact dynamics. The simulations were run using an Intel i7-7700HQ CPU with \SI{16}{\giga\byte} of memory. 

\begin{figure}[!htb]
    \centering
    \includegraphics[width=0.8\linewidth]{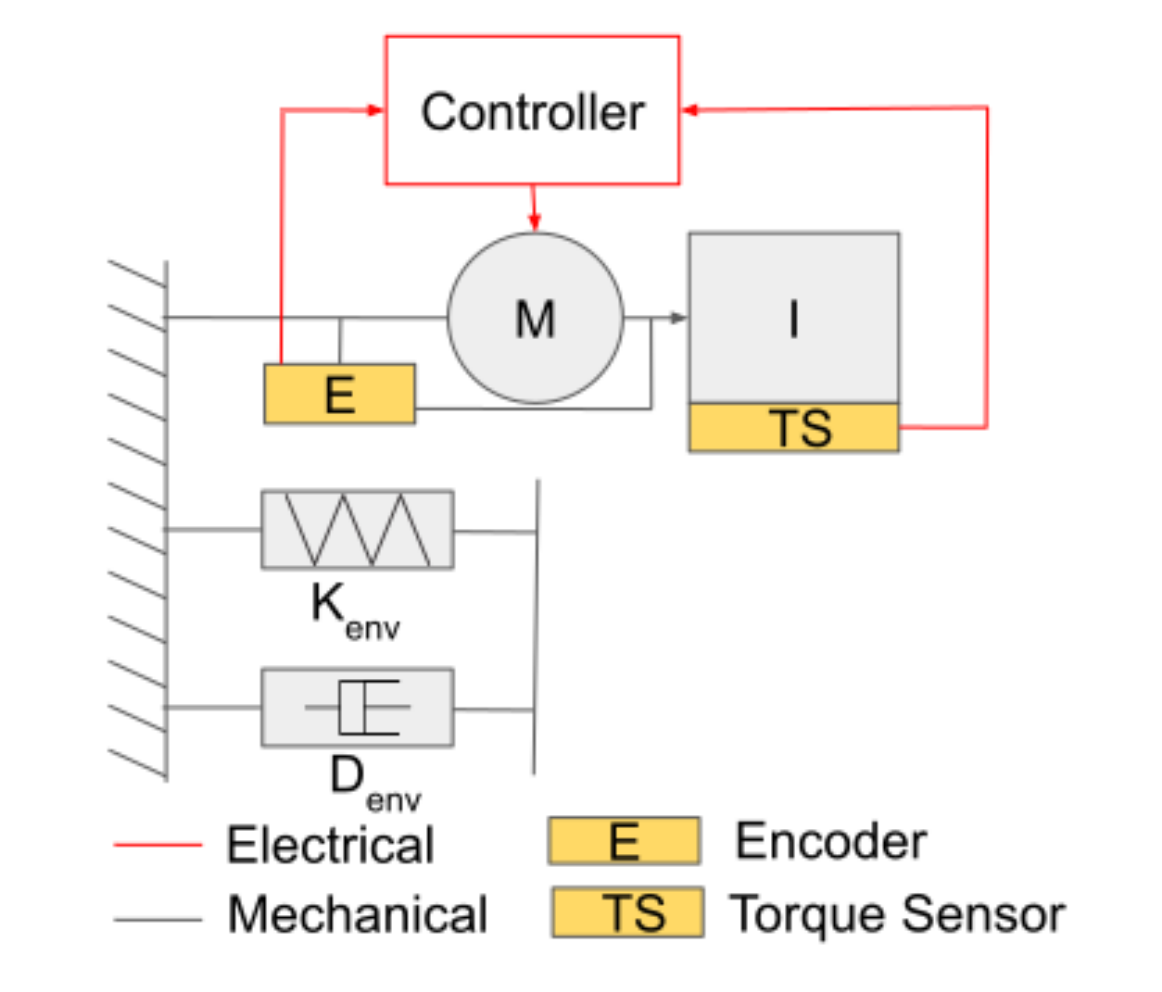}
    \caption{Simulation experimental setup. The motor, inertia and sensors have been implemented using an ideal torque source, inertia, and ideal sensors included in the Simscape library. The spring damper behaviour has been implemented using a rotation hard-stop also present in the Simscape library to implement the contact interaction.}\label{fig:3}
\end{figure}

\begin{figure*}[!htb]
    \centering
    \includegraphics[scale=0.8]{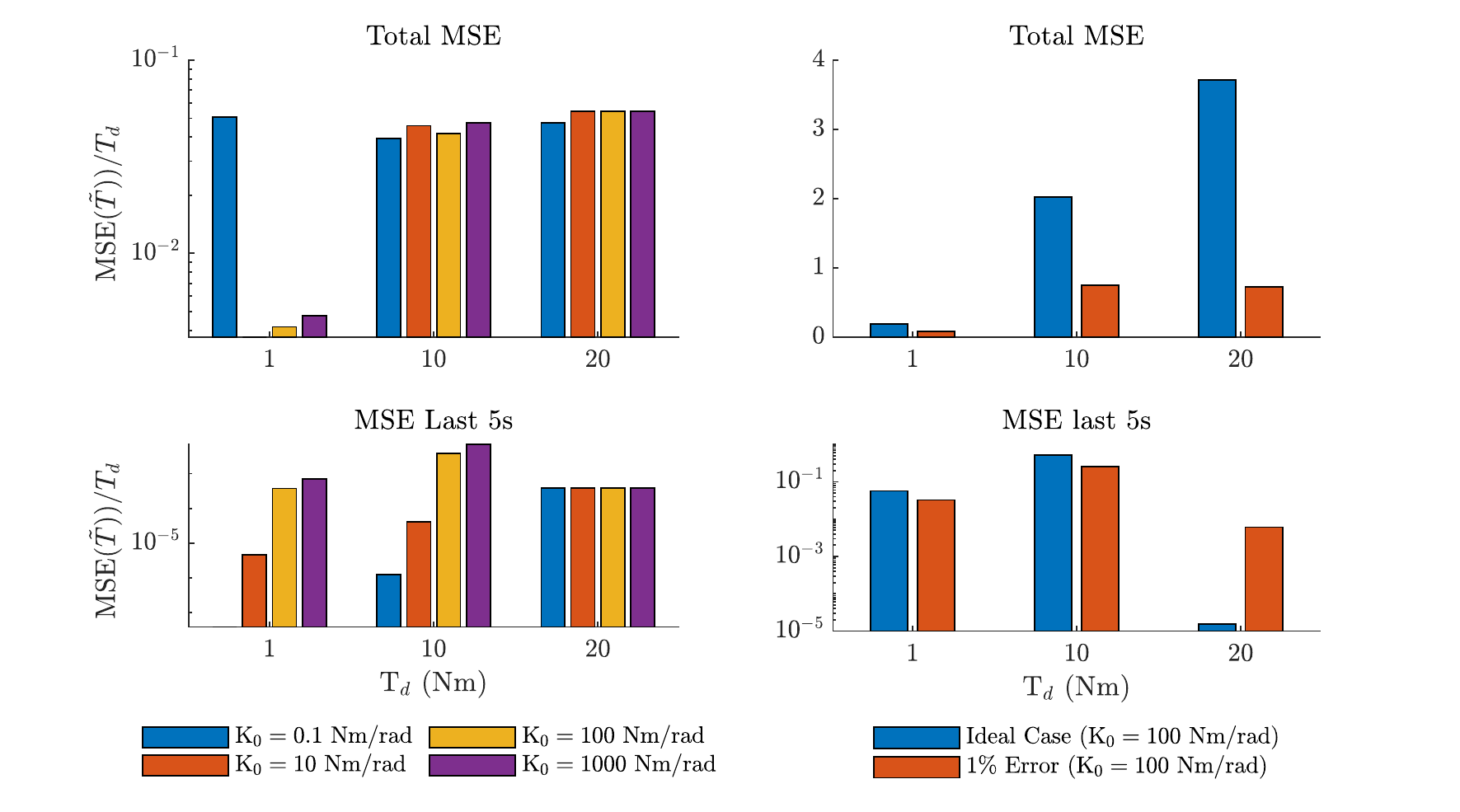}
    \caption{On the left column, there are the normalised Mean Square Error (MSE) during the welded interaction with a spring-damper mechanism. The data show that the error is always negligible regardless of the selected $K_0$. On the right, the normalised MSE for the contact interaction using the same $K_{\text{env}}$ and $D_{\text{env}}$ of the welded interaction show an increase of the error that is probably related to the introduction of the contact dynamics. The \SI{1}{\percent} Error is evaluated setting the $\tilde{x}_d=0.001$ \si{\radian}. Lastly, it shall be noted that the normalisation for T$_\text{d}=20$ \si{\newton} has been performed using F$_{\text{Max}}=15$ \si{\newton}.}\label{fig:4}
\end{figure*}

The simulation setup is described in Fig. \ref{fig:3}, and the system dynamics simulated for \SI{10}{\second}. The parameters that have been kept unchanged across all the simulation are $F_{\text{Max}}=15$ \si{\newton\metre}, $\tilde{x}_{\text{b}}=0.11$ \si{\radian}, $\tilde{x}_{0}=0.10$ \si{\radian} and $I=10$ \si{\kilo\gram\metre\second\squared}. The implemented hard-stop has no transition region for the dynamics and an undamped rebound. The experiments presented in this paper target the study of the influence of the environmental stiffness changes on the controller performance and the evaluation of the effect of contact estimation error on the controller's performance. 

\textit{Welded Parallel Connection To Spring-Damper Mechanism}: The simulation studies a system controlled by the proposed method "welded" to a spring-damper mechanism. The environmental parameters have been set to K$_{\text{env}}=100$ \si{\newton\metre\per\radian} and D$_{\text{env}}=189.7$ \si{\newton\metre\second\per\radian}. The values of K$_0$ chosen for this simulation are $0.1$, $10$, $100$, and $1000$ \si{\newton\metre\per\radian}.
The strongest descriptors of the qualitative behaviour of the system are the relative stiffness and damping between the controller and the environment. The parameters selected for the experiments cover the different emergent behaviours.

\begin{figure*}[!htb]
    \centering
    \includegraphics[scale=0.9]{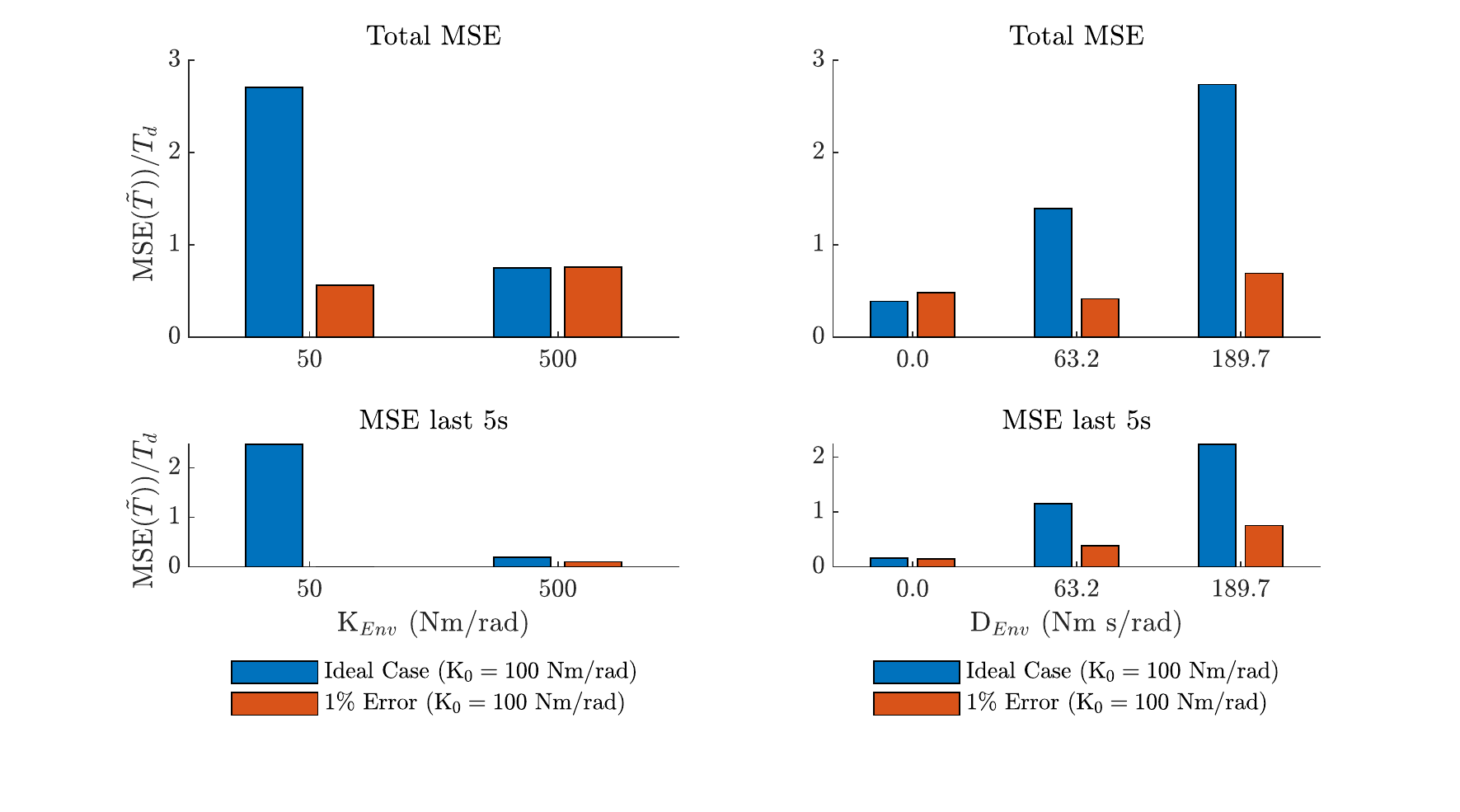}
    \caption{On the left, the normalised MSE is recorded while modifying $K_{\text{env}}$. On the right, the normalised MSE is observed while adjusting $D_{\text{env}}$. In most cases the normalised MSE does not reach acceptable levels even when considering only the final \SI{5}{\second} of the trajectory, indicating that the controller parameters should be tuned to accurately track forces in different environments.}\label{fig:5}
\end{figure*}

\textit{Contact Interaction with a Spring-Damper Mechanism}: The controller constant stiffness selected for these experiments is K$_0=100$ \si{\newton\metre\per\radian} and is equal to the K$_{\text{env}}$ used in the previous experiment. This simulation is subdivided as follows: 
\begin{enumerate}
    \item Effect of the change from welded to contact connection changes the system evolution.
    \item Interaction with different values of environmental stiffness. The (K$_{\text{env}}$) evaluated are \SI{50}{\newton\metre\per\radian} and \SI{500}{\newton\metre\per\radian}.
    \item Performance changes introduced by different (D$_{\text{env}}$) values. The selected values are $0$ and \SI{63.2}{\newton\metre\second\per\radian}, which is the critical damping for a system with $I=10$ \si{\kilo\gram\metre\second\squared} and K$_{\text{env}}=100$ \si{\newton\metre\per\radian}.
\end{enumerate}
During the first simulation, the inertial component starts in contact with the hard-stop, while it starts at \SI{0.5}{\radian} away for the other two simulations. T$_d$ for the last two experiments have been chosen to be \SI{5}{\newton}, which is in the linear stiffness region ($\tilde{x}<\tilde{x}_0$) for the selected parameters. 

\section{Results}
\begin{figure}[!htb]
    \centering
    \includegraphics[scale=0.9]{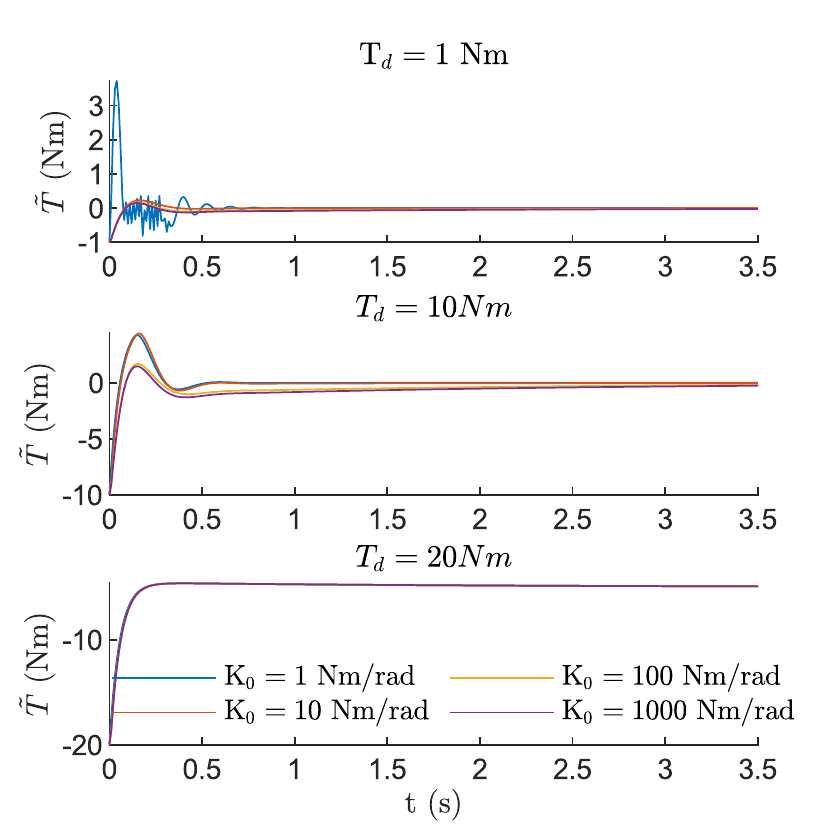}
    \caption{The evolution of the torque error over time for different stiffness shows that the higher normalised MSE for $K_0=1$ and $T_d=1$ is mainly related to a too wide search step implemented by Algorithm \ref{alg2}. It can also be seen that the torque error for $T_d=20$ saturates at \SI{5}{\newton} while the controller saturation torque is set at \SI{15}{\newton}.}\label{fig:6}
\end{figure}

The results for the Normalised Mean Square Error of the torque tracking error ($\tilde{T}$) for the simulations are reported in Fig. \ref{fig:5} and Fig. \ref{fig:6}. The MSEs for the welded connection are always contained below 10$^{-1}$ of the desired torque even when considering the transient period, and they drop below 10$^{-4}$ in the last \SI{5}{\second} of the simulated trajectories. However, the introduction of the contact connection shows an increase in $\tilde{T}$ to about 10$^{-1}$ of the desired torque even at the regime. This indicates that a unilateral connection to an external system is already a source of uncertainty that degrades the performance. The data reported in Fig. \ref{fig:6} further indicate that the errors increase even more when introducing the impact with the external environment, wherein some cases they can even reach two times the desired torque. 

\begin{figure}[!htb]
    \centering
    \includegraphics[scale=0.875]{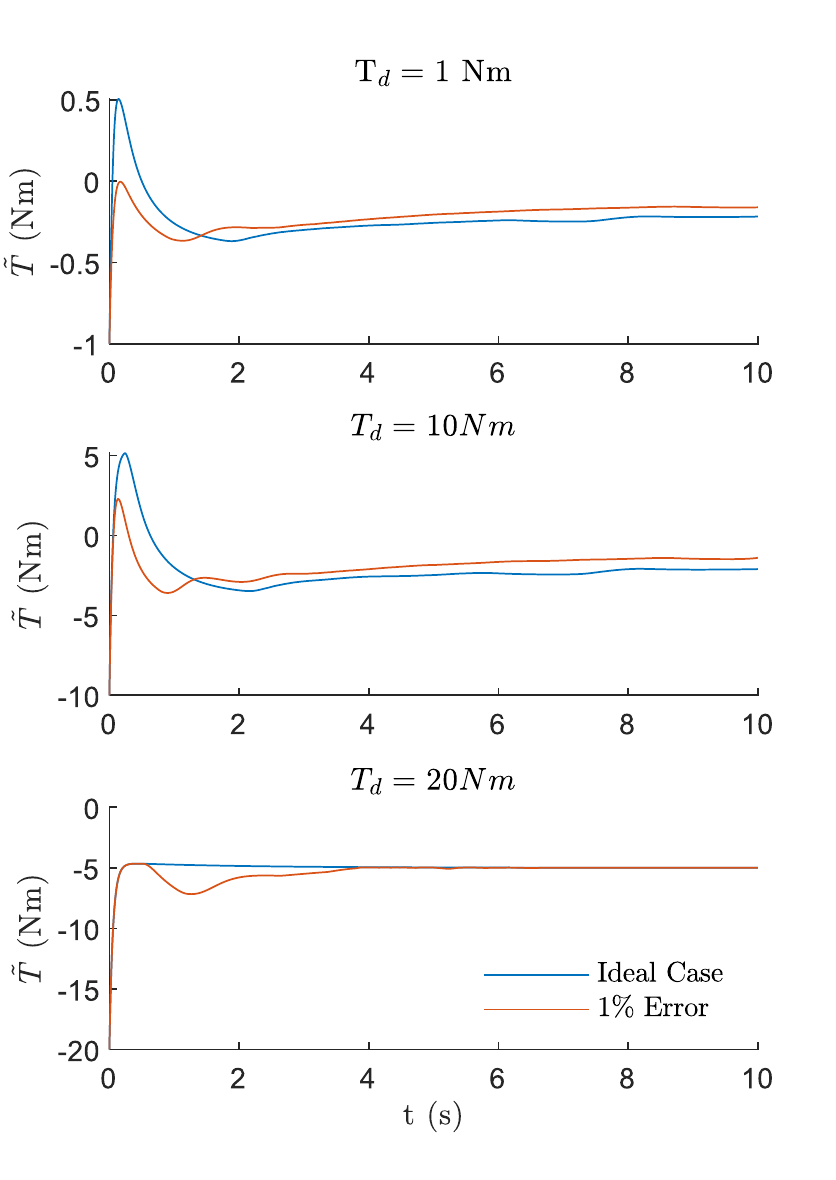}
    \caption{The introduction of the contact dynamics renders the torque evolution smoother for $K_0=1$ and $T_d=1$ and generally increases the convergence time for all the considered cases.}\label{fig:7}
\end{figure}

\begin{figure}[!htb]
    \centering
    \includegraphics[scale=1]{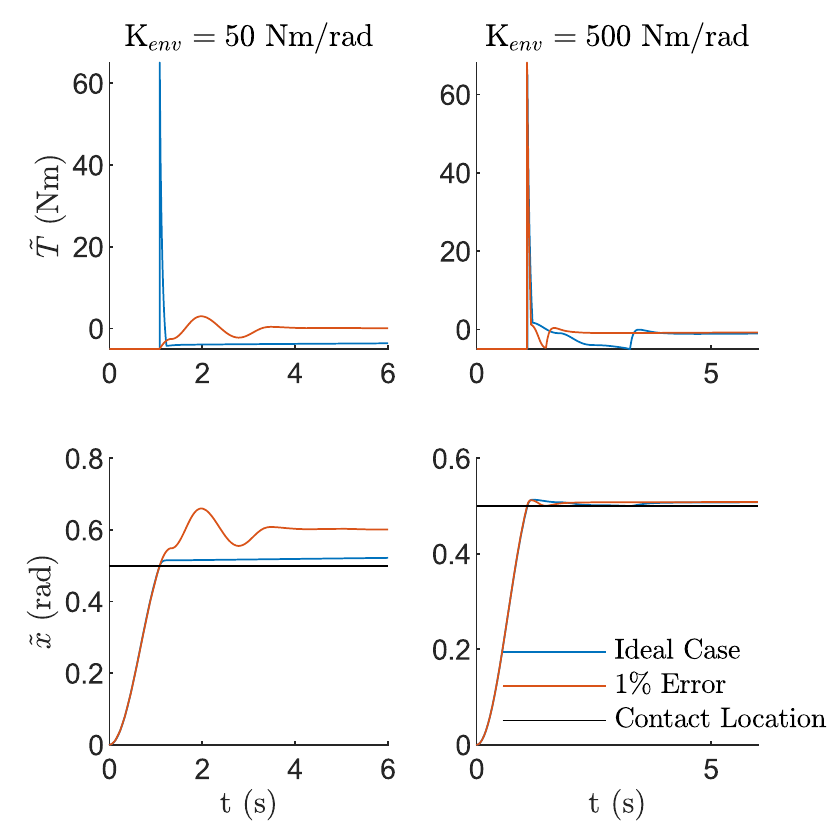}
    \caption{After making contact the system never breaks it indicating the robustness of the controller which can deal with the impact dynamics without any significant issues. The residual error shows that the applied torque is less than the desired. A decreasing trend is still present in all case but with an extremely low rate of convergence.}\label{fig:8}
\end{figure}

\begin{figure}[!htb]
    \centering
    \includegraphics[scale=0.85]{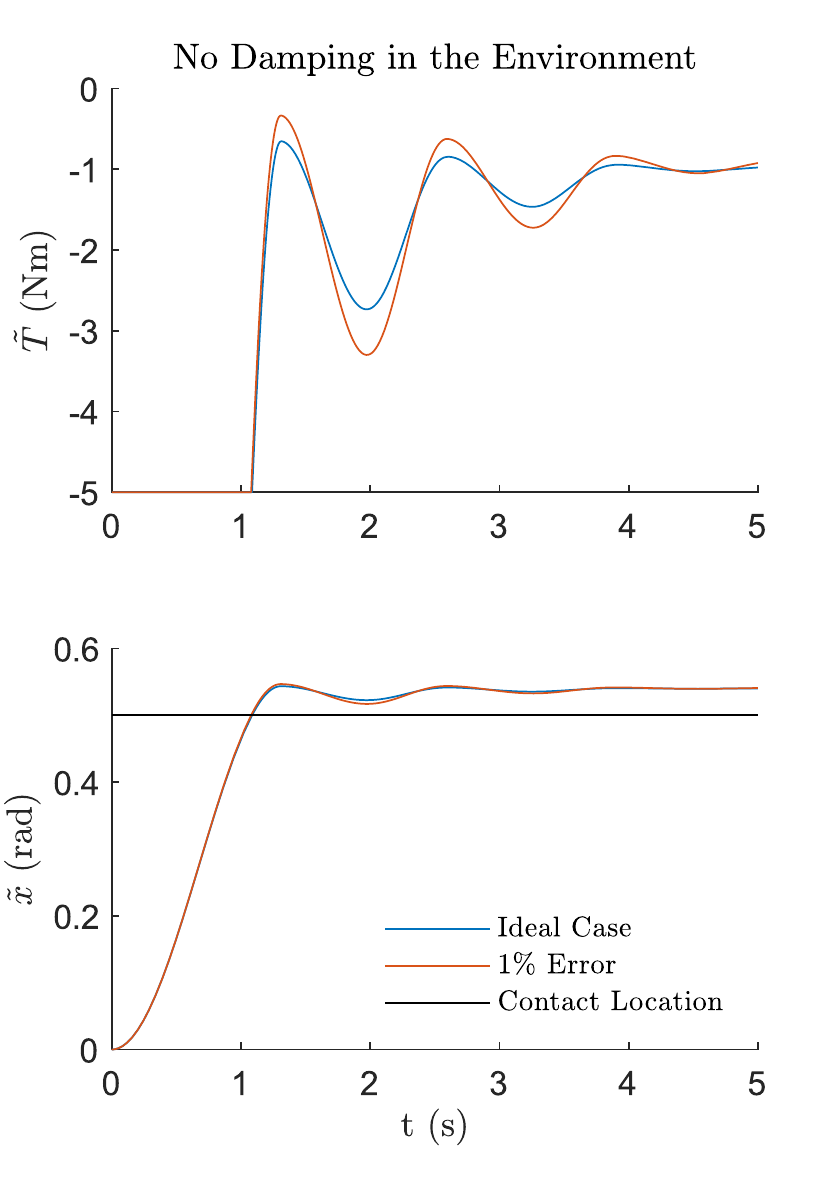}
    \caption{The interaction with an undamped environment shows similar post-contact behaviour of the damped interactions in Fig. \ref{fig:8}. However, the removal of the damping introduces an initial oscillatory behaviour that replaces the torque spike present when making contacts in Fig. \ref{fig:8}.}\label{fig:9}
\end{figure}

The time trajectories of the force and position tracking show a transient of about \SI{0.5}{\second} in the case of welded contact (Fig \ref{fig:6}), which increases to about \SI{1}{\second} when introducing the contact dynamics (Fig \ref{fig:7}). The time data analysis for the other experiments shows that despite the high impulsive force generated at the impact, the controller can retain contact in all the simulations, reported in Fig. \ref{fig:8} and Fig. \ref{fig:9}. Furthermore, the data show that the torque error is always negative (T$<$T$_\text{d}$). $\tilde{T}$ is still reducing but at an extremely slow rate indicating that may be related to the choice of $\sigma$ in Algorithm \ref{alg2}. 


\section{Discussion}
The results indicate that the proposed method can safely interact with external dynamics, and it can deal with the impulsive perturbation received when making contact. The controller passivity also enables to modify the controller parameters online without concerns for its stability.

The data also show that some trade-off needs to be made for obtaining robustness of interaction, particularly when tracking the desired forces. However, other methods are not exempt from similar trade-offs on the torque tracking accuracy when increasing interaction robustness. Differently from these methods, the proposed controller stability is not associated with the external environment but it is an intrinsic property of the controller. Therefore, it requires fewer assumptions on the environmental dynamics, and if they are violated it degrades the system accuracy but it does not lead to a catastrophic failure of the controller.
This inherently safe behaviour is one of the key advantages of our proposed method for the control of real-world robotic systems interacting safely with people and environments.

The integration of Algorithms \ref{alg1} and \ref{alg2} into the architecture presented in Fig. \ref{fig:00} has shown that it is possible to reproduce behaviours similar to the one generated by agonist and antagonist muscles in biological systems. This architecture enables the online adaptation of the impedance behaviour without affecting the system stability, which is similar to what has been observed in human by previous studies \cite{franklin2008,Mussa-Ivaldi1985,todorov2000,todorov2002}. The ability to retain the stability during a wide range of unknown dynamic conditions is desirable for applications such as human-robot interaction (e.g., haptics, rehabilitation robots, and prosthetics), control of cable-driven systems where it may difficult to accurately model the transmission dynamics, and soft robotics. This framework may also find application in modelling human motor control, where the possibility of superimposing multiple controllers may provide a platform to better understand motor synergies.

In conclusion, the proposed method has been proven feasible, and it shows a good level of performance when it is applied on a bilateral (welded) connection to the environment. The introduction of unilateral contact dynamics increases the force tracking error, however, it does not interfere with the ability of the controller to retain contact and stability of interaction. The residual tracking error seems to be related to the choice of the $\sigma$ parameter in Algorithm \ref{alg2}, which should be exposed and optimised by interaction. However, the data also show that this problem can be solved simply by setting $F_{\text{Max}}=T_d$. This also enables a higher convergence speed to the desired torque. The passivity of the controller further implies that multiple controllers can be superimposed without interfering with the system stability.
This provides a framework to develop a hierarchical architecture of semi-autonomous controllers that can be used to study human motor control and improve robots dynamic interaction performances---a focus of our future work.

\section*{Appendix}
\subsection{Lyapunov's Stability Analysis} \label{sec:stabilityAnalysis}
The fractal impedance has a non-smooth piece-wise energy manifold with a time-invariant topology that scales with the controller gains by it does not change shape \cite{babarahmati2019}. Let's now consider the controller autonomous dynamics for a monodimensional system generated via Algorithm \ref{alg1}:
		\begin{equation}
	\label{Eq:dynamicdot}
	   \begin{cases}
	    \Lambda\ddot{x}  + F_\text{K}(\tilde{x})=0\\
	    \Lambda\ddot{x}  + F_\text{AA}(\tilde{x})=0
	    \end{cases}
	\end{equation}
	
	\noindent where $\Lambda$ is the inertia of the system. A valid Lyapunov's candidate is:
	\begin{equation}
	\label{eq:LyapunvFunction}
	V=\begin{cases}
	V_{\text{K}}= (\dot{x}^{T} \Lambda \dot{x})/2 +E_\text{K}(\tilde{x})\\
	V_{\text{AA}}= (\dot{x}^T\Lambda \dot{x})/2 +E_\text{AA}(\tilde{x}) + E_{\text{C}}
	\end{cases}
	\end{equation}
\noindent where $E_{\text{C}}$ is a constant of energy offset introduced in the switching conditions. Thus, V time derivative is:
	\begin{equation}
	\label{Eq:Vdot}
	   \begin{cases}
	    \dot{V}_\text{K}=(\Lambda\ddot{x}  + F{K}(\tilde{x}))\dot{x}=0\\
	    \dot{V}_\text{AA}=(\Lambda \ddot{x} + F_{AA}(\tilde{x}))\dot{x}=0
	    \end{cases}
	\end{equation}
  
 Therefore, the conditions for stability are respected in both the branches of the system. Nevertheless, being the system non-smooth to prove stability is needed to verify if  V is a Lipschitz function during the switching conditions. Being $\dot{V}=0$ at the switching conditions due to $\dot{x}=0$. 
 \begin{equation}
	\label{eq:LyapunvFunction2}
	\begin{cases}
	\displaystyle{\lim_{\substack{\tilde{x}\to \tilde{x}_{Max} \\ \dot{x}\to0 ~~~}}{V_{\text{K}}}=E_{\text{K}}(\tilde{x}_{Max})}\\
    \displaystyle{\lim_{\substack{\tilde{x}\to \tilde{x}_{Max} \\ \dot{x}\to 0 ~~~}}{V_{\text{AA}}}=E_{\text{AA}}(\tilde{x}_{Max})+E_{\text{C}}}\\
	\displaystyle{\lim_{\substack{\tilde{x}\to0~~~ \\\dot{x}\to0 ~~~}}{V_{\text{K}}}=E_{\text{K}}(0)=0}\\
    \displaystyle{\lim_{\substack{\tilde{x}\to0~~~ \\\dot{x}\to0 ~~~}}{V_{\text{AA}}}=E_{\text{AA}}(0)+E_{\text{C}}}\\
	\end{cases}
\end{equation}
The continuity condition can be derived using the following system of equations:
 \begin{equation}
	\label{eq:LyapunvFunction3}
	\begin{cases}
	E_{\text{C}}=E_{\text{K}}(\tilde{x}_{Max}) -E_{\text{AA}}(\tilde{x}_{Max})\\
	E_{\text{C}}=-E_{\text{AA}}(0)\\
	\end{cases}
\end{equation}
being:
\begin{equation*}
    	\label{eq:LyapunvFunction4}
    	\left\{\begin{array}{ll}
    	E_{\text{AA}}(0)& =\displaystyle{\int_{\tilde{x}_{Max}/2}^0F_{\text{AA}}(\tilde{x})d\tilde{x}}=\displaystyle{-\int^{\tilde{x}_{Max}/2}_0F_{\text{AA}}(\tilde{x})d\tilde{x}}\\\\&
    	= -(E_{\text{K}}(\tilde{x}_{Max})-\Delta E_{\text{A}|\tilde{x}_{Max}})/2 \\\\
    	E_{\text{AA}}(\tilde{x}_{Max})&=(E_{\text{K}}(\tilde{x}_{Max})+\Delta E_{\text{A}|\tilde{x}_{Max}})/2
    	\end{array}\right.
\end{equation*}

\noindent and given that the energy contribution of antagonist muscle ($\Delta E_{\text{A}|\tilde{x}_{Max}}$) always opposes to the motion:
\begin{equation}
	\label{eq:LyapunvFunction5}
	E_{\text{C}}=-E_{\text{AA}}(0)=(E_{\text{K}}(\tilde{x}_{Max})-\Delta E_{\text{A}|\tilde{x}_{Max}})/2
\end{equation}	
which implies:
\begin{equation}
	\label{eq:LyapunvFunction6}
\begin{cases}
	V_{\text{AA}}(0)=0 \\
    V_{\text{AA}}(\tilde{x}_{Max})=E_{\text{K}}(\tilde{x}_{Max})
\end{cases}
\end{equation}
%%%%%%%%%%%%%%%%%%%%%%%%%%%%%%%%%%%%%%%%%%%%%%%%%%%%%%%%%%%%%%%%%%%%%%%%%%%%%%%%
\bibliography{root}
\bibliographystyle{IEEEtran}
\end{document}